\documentclass[runningheads, dvipsnames]{llncs}

\usepackage{amssymb}
\usepackage{booktabs}
\usepackage{multirow}
\usepackage[inline]{enumitem}

\usepackage{xcolor}
\usepackage{graphicx}
\let\llncssubparagraph\subparagraph
\let\subparagraph\paragraph
\usepackage[compact]{titlesec}
\let\subparagraph\llncssubparagraph

\titlespacing*\section{0pt}{12pt plus 4pt minus 2pt}{2pt plus 2pt minus 2pt}
\titlespacing*\subsection{0pt}{12pt plus 4pt minus 2pt}{2pt plus 2pt minus 2pt}
\titlespacing*\subsubsection{0pt}{12pt plus 4pt minus 2pt}{2pt plus 2pt minus 2pt}

\begin{document}
\title{Variance Reduction in Ratio Metrics for Efficient Online Experiments}
%
%

\author{Shubham Baweja, Neeti Pokharna, Aleksei Ustimenko and Olivier Jeunen}

%
%
\institute{ShareChat}
\maketitle              
\begin{abstract}
Online controlled experiments, such as A/B-tests, are commonly used by modern tech companies to enable continuous system improvements.
Despite their paramount importance, A/B-tests are expensive: by their very definition, a percentage of traffic is assigned an inferior system variant.
To ensure statistical significance on top-level metrics, online experiments typically run for several weeks.
Even then, a considerable amount of experiments will lead to \emph{inconclusive} results (i.e. false negatives, or type-II error).
The main culprit for this inefficiency is the variance of the online metrics.
Variance reduction techniques have been proposed in the literature, but their direct applicability to commonly used ratio metrics (e.g. click-through rate or user retention) is limited.

In this work, we successfully apply variance reduction techniques to ratio metrics on a large-scale short-video platform: ShareChat.
Our empirical results show that we can either improve A/B-test confidence in 77\% of cases, or can retain the same level of confidence with 30\% fewer data points.
Importantly, we show that the common approach of including as many covariates as possible in regression is counter-productive, highlighting that control variates based on Gradient-Boosted Decision Tree predictors are most effective.
We discuss the practicalities of implementing these methods at scale and showcase the cost reduction they beget.
\end{abstract}


\section{Introduction \& Motivation}
Online controlled experiments (colloquially known as A/B-tests) are a crucial tool for online businesses to enable data-driven decision making~\cite{kohavi2020trustworthy,Larsen2023}. 
Nevertheless, they are inherently costly and statistical nuances pose challenges when implementing them at scale and interpreting their results correctly~\cite{Jeunen2023_misassumption,Kohavi2022}.

The motivation behind this research is to enhance the confidence and efficiency of A/B-testing, particularly when dealing with commonly used \emph{ratio} metrics such as click-through rates and user retention.
These metrics play a crucial role in evaluating user engagement and platform performance but inhibit the use of classical statistical tools.
Our goal is to increase the \emph{sensitivity} of such ratio metrics.
This can either lead to
\begin{enumerate*}[label=(\roman*)]
    \item more conclusive results at constant sample sizes (i.e. reduced type-II error), or
    \item a reduction of the required sample size to obtain constant $p$-values (i.e. saving time and experiment cost).
\end{enumerate*}

Variance reduction techniques have been proposed in the literature~\cite{Budylin2018,Deng2013,Guo2021,Poyarkov2016}, but their direct applicability to ratio metrics is not always straightforward.
Furthermore, a myopic focus on \emph{variance} can make us blind to \emph{bias}~\cite{minasyan2021estimation}, implying that \emph{variance} reduction does not unequivocally lead to \emph{sensitivity} improvements.

We discuss the practical application of a synthesis of existing variance reduction techniques to improve sensitivity (or, analogously, decrease experimentation cost) on the ShareChat platform.
Our empirical observations, based on a log of historical real-world A/B-experiments, underline our theoretical insights.
We find that the highest variance reduction does not map to the highest sensitivity increase.
Our best-performing method can increase statistical confidence in 77\% of A/B-experiments, or achieve on-par confidence with 30\% fewer samples, directly impacting the cost per experiment for the business.
\section{Problem Statement, Methodology \& Contribution}
\subsection{Statistical Hypothesis Testing}

Consider an online controlled experiment with deployed variants $A$ and $B$, and a metric $\mathcal{M}$ to evaluate.
That is, $\mathcal{M}$ is a random variable whose expectation we wish to estimate under $A$ and $B$.
Typically, a $z$-test is done to ascertain whether the difference in observed metric values is statistically significant.
Denote by $\mu(\mathcal{M})$ the mean of the observed values, and by $\sigma(\mathcal{M})$ its standard deviation.
A subscript $\mu_{V}$ indicates the mean over observed values for units assigned to $V$.

The $z$-statistic is then given by Eq.~\ref{eq:zscore}, which can be transformed to a two-tailed $p$-value for the null hypothesis $A\simeq B$ following Eq.~\ref{eq:pvalue}, where $\Phi(\cdot)$ represents the cumulative distribution function (CDF) for a standard Gaussian.
\noindent\begin{minipage}{0.49\linewidth}
\begin{equation}\label{eq:zscore}
    z_{A\succ B}(\mathcal{M}) = \frac{\mu_{A}(\mathcal{M}) - \mu_{B}(\mathcal{M})}{\sqrt{\frac{\sigma_{A}(\mathcal{M})^{2}}{N_{A}} + \frac{\sigma_{B}(\mathcal{M})^{2}}{{N_{B}}}}}
\end{equation}
\end{minipage}%
\begin{minipage}{0.51\linewidth}
\vspace{-3ex}
\begin{equation}\label{eq:pvalue}
    \quad
    p_{A\succ B}(\mathcal{M})  = 2\Phi\left(-\big|z_{A \succ B}(\mathcal{M})\big|\right)
\end{equation}
\end{minipage}

For a confidence level of $\alpha$ (typically $\approx 0.05$), we can reject the null hypothesis when $p_{A\succ B}(\mathcal{M}) < \alpha$, and claim a statistically significant impact on metric $\mathcal{M}$.
Many standard metrics (e.g. daily active users, event counters, et cetera) can be expressed as Bernoulli-distributed random variables.
In these cases, $\mu(\mathcal{M})$ and $\sigma(\mathcal{M})$ can be straightforwardly computed using standard formulas.
Aside from these metrics, modern platforms on the web typically also care about \emph{ratio} metrics (e.g. click-through rate, retained users per active users, et cetera).
In these cases, we need to apply the Delta method to estimate the variance in the denominator of Eq.~\ref{eq:zscore}.
For a metric $\mathcal{M} \equiv \frac{\mathcal{M}_{N}}{\mathcal{M}_{D}}$, this yields:
\begin{equation}\label{eq:delta_method}
\sigma(\mathcal{M})^{2} \approx \frac{\mu(\mathcal{M}_{N})^{2}}{\mu(\mathcal{M}_{D})^{2}}
\left(
    \frac{\sigma(\mathcal{M}_{N})^{2}}{\mu(\mathcal{M}_{N})^{2}} + \frac{\sigma(\mathcal{M}_{D})^{2}}{\mu(\mathcal{M}_{D})^{2}} - 2 \frac{\mathrm{cov}(\mathcal{M}_{N},\mathcal{M}_{D})}{\mu(\mathcal{M}_{N})\mu(\mathcal{M}_{D})}.
\right)
\end{equation}

Alternatively, one can adopt a linearisation approach to obtain a new metric $L(\mathcal{M}) = \mathcal{M}_{N} - c\mathcal{M}_{D}$ that preserves directionality and statistical power~\cite{Budylin2018}.
Here, $c = \frac{\mu_{C}(\mathcal{M}_{N})}{\mu_{C}(\mathcal{M}_{D})}$, where $C$ corresponds to the \emph{control} variant.

\subsection{Variance Reduction in Online Controlled Experiments}\label{sec:cuped}
Lower $p$-values (i.e. higher $z$-scores) indicate higher confidence in rejecting the null hypothesis.
From Eq.~\ref{eq:zscore}, a clear way to increase the $z$-score is to increase the sample size $N$.
As this corresponds to running the experiment for a longer period or on a larger portion of traffic, this is costly.
An alternative route is to instead decrease the variance of the metric $\sigma(\mathcal{M})^{2}$.
This can be done by leveraging \emph{control variates}.
Suppose we have access to a random variable $\mathcal{M}_{\rm CV}$ that is correlated with the metric value $\mathcal{M}$, but \emph{independent} of the treatment.
Then, we can obtain a variance-reduced metric simply by subtracting it:
\begin{equation}\label{eq:cv}
    \mathcal{M}_{\rm VR} = \mathcal{M} - \mathcal{M}_{\rm CV}.
\end{equation}
The CUPED approach proposes to assign pre-experiment values of the same metric $\mathcal{M}_{\rm pre}$ as the control variate in a linear regression model~\cite{Deng2013}:
\begin{equation}
    \mathcal{M}_{\rm CV} = \theta(\mathcal{M}_{\rm pre} - \overline{\mathcal{M}_{\rm pre}}), \rm{~with~} \theta = \frac{{\rm cov}(\mathcal{M}, \mathcal{M}_{\rm pre})}{\sigma(\mathcal{M}_{\rm pre})^{2}} \rm{~estimated~on~pooled~data}.
\end{equation}
As a natural extension, multiple covariates can be used in a regression model to obtain a final estimate for the outcome metric.
The variance reduction this approach yields is directly proportional to the correlation between the original metric $\mathcal{M}$ and the control variate $\mathcal{M}_{\rm CV}$.
Minasayan et al. provide a characterisation of the approximation bias on the Average Treatment Effect (ATE) that is incurred by this approach~\cite{minasyan2021estimation}, which is multiplicative of the form $1-\frac{k}{N}$ with $k$ the number of regression covariates.
This implies that for fixed $N$, adding more covariates always leads to a \emph{decrease} in the ATE estimate.
This, in turn, leads to a \emph{decrease} in the \emph{numerator} of Eq.~\ref{eq:zscore}, which might (partially) offset variance reduction: lower variance does not unequivocally imply higher sensitivity.

Sidestepping the problems of approximation with linear models, Poyarkov et al. propose to use Gradient-Boosted Decision Trees (GBDTs) to estimate $\mathcal{M}_{\rm CV}$ directly, reporting significant improvements over linear models and CUPED~\cite{Poyarkov2016}.

\subsection{Variance Reduction for Ratio Metrics}
We train a multiple linear regression model to estimate the ATE on a ratio metric $\mathcal{M}$ from past online experiments, based on multiple covariates.
Following the classical CUPED method~\cite{Deng2013} extended for ratio metrics, the covariates include: the numerator from the pre-experiment period $\mathcal{M}_{N, {\rm pre}}$; the denominator from the pre-experiment period $\mathcal{M}_{D, {\rm pre}}$; and the linearised metric for the pre-experiment period $L(\mathcal{M}_{\rm pre})$ (following~\cite{Budylin2018}).
Alternatively, we consider GBDT predictions for the numerator, denominator, and linearisation $\widehat{\mathcal{M}_{N}}, \widehat{\mathcal{M}_{D}}, \widehat{L(\mathcal{M})}$ (following~\cite{Poyarkov2016}).
The output of the regression model is then the control variate that we plug into Eq.~\ref{eq:cv} to obtain a variance-reduced metric, on which we perform a $z$-test following Eqs.~\ref{eq:zscore} and \ref{eq:pvalue}.
\section{Experimental Validation of Improved Sensitivity}
\begin{table*}[t!]
\vspace{-2ex}
\centering
\caption{Empirical insights into variance-reduced metrics on a dataset of past A/B experiments.
We consider using pre-experiment metrics $\mathcal{M}_{\rm pre}$, GBDT-based predictions $\widehat{\mathcal{M}}$, or both.
Lower variance does not unequivocally imply lower $p$-values, and GBDT-based predictions lead to significantly improved sensitivity with good type-I errors.}
{\vspace{-2ex}
\begin{tabular}{lcrcrcrcr}
\toprule
\textbf{Covariates} &~~~& \textbf{Var. Red.} &~~~& \textbf{$\mathbb{P}(p\rm{-value}\downarrow)$}&~~~& \textbf{med. rel. $z$} &~~~& \textbf{Type-I Error}\\
\midrule
$\mathcal{M}_{\rm pre}$ &~& -72.47~\% &~& 15.38~\% &~& 0.78 &~& 4.3\%\\
$\widehat{\mathcal{M}}$  &~& -45.66~\% &~& \textbf{76.92}~\% &~& \textbf{1.19} &~& 4.8\% \\
$ \{\mathcal{M}_{\rm pre}\} \cup \{\widehat{\mathcal{M}}\}  $ &~& -\textbf{72.62}~\% &~& 30.77~\% &~& 0.81 &~& 5.2\% \\
\bottomrule
\end{tabular}
}
\label{tab:results}\vspace{-5ex}
\end{table*}
We consider past online controlled experiments that ran on the ShareChat application, with known (conclusive) outcomes.
That is, for deployed system variants $A$ and $B$, we know a preference ordering to obtain $A \succ B$.
All experiments ran for either 7 or 14 days, and all variants were assigned between 1.6 and 3.5 million users.
In this preliminary study, we consider 13 known A/B pairs. 
The ratio metric we consider is \emph{one-day retention}: the number of users who were retained from day $D_{0}$ to $D_{1}$, over the number of active users on day $D_{0}$, aggregated over the experiment period.
For this metric and the variance reduction approaches introduced earlier, we measure several attributes:
\begin{enumerate*}[label=(\roman*)]
    \item the reduction in variance compared to the original ratio metric $\mathcal{M}$,
    \item the fraction of online experiments that have a lower $p$-value under the variance-reduced metric,
    \item the median relative $z$-score, which gives an indication of the sample size reduction that would be gained at constant sensitivity, and
    \item the type-I error measured over 220 two-week A/A-pairs on 3.5 million users per variant.
\end{enumerate*}
Table~\ref{tab:results} presents these results.
Even though pre-experiment metric values contribute significantly to variance reduction, they also significantly impact the ATE estimate, which leads to disappointing performance when considering the percentage of experiments for which we can reject the null hypothesis with higher confidence.
Indeed, when considering solely GBDT-based predictions $\widehat{\mathcal{M}}$, we obtain lower $p$-values for 77\% of experiments.
For the classical CUPED approach using pre-experiment values, this number deteriorates to 15\%.
Combining all covariates does not provide solace, stagnating at 31\%.
This is a direct result of the bias we discuss in Section~\ref{sec:cuped}.
Furthermore, a median relative $z$-score of $1.19$ implies that we need $1.19^{2}\approx 1.42$ times fewer data points to achieve the same level of confidence as the original ratio metric ($1-\frac{1}{1.42}\approx 30\%$; see e.g.~\cite{Chapelle2012,Kharitonov2017}).
This directly impacts the speed at which experiments conclude, and the number of experiments that can run concurrently, greatly improving experimental velocity.
For $\alpha=0.05$, we expect the type-I errors to converge at 5\% as well.
We observe that all are very close, ensuring that our encouraging results are not the result of overfitting.

\section{Conclusions \& Outlook}
In this work, we have explored variance reduction techniques specifically tailored for ratio metrics in the context of A/B-testing. 
One notable takeaway from our study is a re-evaluation of the conventional CUPED method~\cite{Deng2013}.
This approach, which often incorporates multiple covariates (pre-experiment metrics) can inadvertently lead to overfitting. Instead, we propose using GBDT predictors which are \emph{unbiased} estimators and account for complex relationships among pre-experiment metrics~\cite{Poyarkov2016}.
Leveraging predicted metrics as covariates in the CUPED framework yields lower variance reduction but better performance in terms of sensitivity without inflating type-I errors, providing empirical support for the inclusion of unbiased estimators to allow for more efficient A/B-testing.

\section*{About the presenter}
Aleksei Ustimenko  is a staff applied scientist at ShareChat, focusing on A/B experimentation, statistics and recommender systems.
He additionally enjoys work as a theoretical mathematician, focusing on stochastic calculus.

\section*{About the company}
ShareChat (Mohalla Tech Pvt Ltd) is India’s largest homegrown social media company, with 325\textsuperscript{+} million Monthly Active Users (MAUs) across all its platforms, and social media brands such as ShareChat App and Moj under its portfolio.
Today, ShareChat App is India’s leading social media platform with over 180 million MAUs spread across the country, and Moj is India’s \#1 short-video app with the highest MAU base of nearly 160 million.

%
%
%
\bibliographystyle{splncs04}
\bibliography{references}

\end{document}